\documentclass{article}

\PassOptionsToPackage{numbers,compress}{natbib}
\usepackage[final,main]{neurips_2026}

\usepackage[utf8]{inputenc}
\usepackage[T1]{fontenc}
\usepackage{hyperref}
\usepackage{url}
\usepackage{booktabs}
\usepackage{amsfonts}
\usepackage{microtype}
\usepackage{xcolor}
\usepackage{amsmath, amssymb, amsthm}
\usepackage{graphicx}
\usepackage{multirow}
\usepackage{subcaption}
\usepackage{array}
\usepackage{wrapfig}
\usepackage{float}
\usepackage{tikz}
\usetikzlibrary{arrows.meta,positioning}

\definecolor{handbg}{HTML}{FFF7E6}
\definecolor{handframe}{HTML}{B7791F}
\definecolor{autobg}{HTML}{EAF4FF}
\definecolor{autoframe}{HTML}{2B6CB0}
\definecolor{pipelineblue}{HTML}{2B6CB0}
\definecolor{pipelinegreen}{HTML}{2F855A}
\definecolor{pipelineorange}{HTML}{B7791F}
\definecolor{pipelinered}{HTML}{C53030}

\title{Automating SKILL.md Generation for Computer-Using Agents via Interaction Trajectory Mining}

\author{
    \textbf{Yuexing Hao}$^{1}$, 
    \textbf{Xiaomin Li}$^{2}$\\
    $^1$Massachusetts Institute of Technology, $^2$Harvard University\\
    \texttt{yuexing@mit.edu}
}

\begin{document}

\maketitle

\begin{abstract}
Explicit skill libraries make computer-using agents easier to inspect, but it remains unclear whether such libraries can be mined from interaction data in a way that improves downstream policies. We study this question through a three-stage pipeline that segments GUI trajectories, clusters segments into candidate skills, and trains a skill-aware policy from the resulting annotations. The mined clusters are readable on the source benchmark: five of eight clusters have at least 0.95 purity against InteraSkill Workflows labels. However, readability does not imply transfer. GRPO improves IW skill-step accuracy only from 18.5\% to 20.5\%, leaves BrowseComp+ essentially unchanged, and underperforms trivial frequency priors on key source-domain metrics. We therefore present the method as a diagnostic study: trajectory mining can expose inspectable skill structure, but the current boundary detector, orderless segment representation, and offline reward model are insufficient for reliable cross-domain policy improvement.
\end{abstract}

\section{Introduction}

Computer-using agents (CUAs) act on graphical user interfaces (GUIs) by clicking, typing, scrolling, copying, and pasting~\cite{yao2022webshop,deng2023mind2web,zhou2024webarena}. As these agents move from single-step web tasks to longer workflows, repeated patterns become important. A user may need to search for a page, copy a value, switch applications, fill a form, or send a message. Agent systems often package these repeated routines as skills: named procedures that sit above primitive user-interface (UI) actions and below full task plans. We use \texttt{SKILL.md} to refer to this kind of explicit skill file. These files make behavior easier to inspect and debug, but they are usually written by hand.

Hand-written skills create a practical bottleneck. They must be named, scoped, documented, and updated as interfaces change. They also encode designer assumptions about which behaviors are reusable. A trajectory dataset contains another possible source of structure: if many users repeatedly perform similar action subsequences, those subsequences may reveal natural skill units. This paper asks whether we can build explicit skill libraries from such trajectories. The hard part is not merely clustering trajectories. The hard part is showing that the discovered clusters help a policy on new tasks. A cluster can be coherent without being useful. A policy can improve because it sees more action data, not because it learned a transferable skill vocabulary. Group Relative Policy Optimization (GRPO) can also optimize the reward model without improving benchmark accuracy. We therefore ask a narrow question: \emph{what transfers from mined skills?}


Our pipeline has three steps. First, we cut trajectories at large action changes. Second, we cluster the resulting segments and refine the cluster embedding with pseudo-label contrastive learning. Third, we train Qwen3-8B with GRPO from the base model. The reward model scores full skill-aware responses, but the main reported Phase~3 evaluation is skill-sequence composition: predicting the next mined skill labels and comparing the resulting skill sequence with the reference sequence. We compare this policy to zero-shot baselines with the same skill-oriented format on IW, WebArena, and BrowseComp+. We additionally report WorkArena-NLP as a text-only diagnostic and a Mind2Web zero-shot baseline as context, but neither is used to claim current GRPO transfer.

The main results are mixed. On the source IW benchmark, the mined library is readable: five of eight clusters have at least 0.95 purity against IW ground-truth skills. As a downstream training signal, however, the current setup is weak. GRPO raises IW skill-step accuracy only from 18.5\% to 20.5\%, while BrowseComp+ skill-step accuracy changes from 43.5\% to 43.3\%. A trivial Frequency baseline is a stronger next-skill predictor on IW than the proposed multi-layer perceptron (MLP) and GRPO, and it has lower edit distance than Auto-\texttt{SKILL.md} at every data size. We therefore treat the result as evidence about the limits of the current pipeline. Trajectory mining can produce readable skill structure. Our current reward model, orderless segment representation, and GRPO setup do not turn that structure into a strong skill-composition policy, and several learned variants underperform a most-common-skill baseline. 

This paper is a diagnostic study rather than a success claim. We make three contributions. First, we present a simple pipeline for mining explicit SKILL.md-style routines from GUI trajectories and show that it produces readable source-domain structure. Second, we evaluate whether these mined skills improve downstream skill composition under several baselines and transfer checks. Third, we report a negative result: the current learned components do not outperform trivial frequency priors, and verified cross-domain gains are absent or negative. These findings clarify which parts of trajectory-mined skill libraries are currently useful, and which parts remain unsolved. Our reproducible codes are available in the anonymous Github repository \footnote{Anonymous project repository: \url{https://anonymous.4open.science/r/CUA-1680}.}.

\section{Related Work}

Modern CUAs use structured UI observations and fixed action spaces. WebShop~\cite{yao2022webshop} and Mind2Web~\cite{deng2023mind2web} define common primitives, while WebArena~\cite{zhou2024webarena}, VisualWebArena~\cite{koh2024visualwebarena}, WorkArena~\cite{drouin2024workarena}, and OSWorld~\cite{xie2024osworld} test realistic web and operating-system tasks. Foundation CUA systems such as OpAgent~\cite{guo2026opagent}, OpenCUA~\cite{wang2025opencua}, and UltraCUA~\cite{yang2025ultracua} push this direction with larger models and trajectory corpora. Our emphasis on inspectable skill files also connects to human-centered and rule-grounded AI work outside GUI-agent benchmarks. Hao et al. study AI systems for shared decision-making with older adult cancer patients~\cite{hao2024shared}, EHR-integrated LLM agents for prostate-cancer patient education~\cite{hao2025mededuchat}, and physician--AI relevance alignment in medical question answering~\cite{hao2025medpair}; Li et al. study guideline-following medical decisions, rule-based data selection, and adaptive safety rules for reward modeling~\cite{li2025medguide,li2024ruledata,li2025safetyrules}. These systems motivate the same design principle we adopt here: automated agents should expose intermediate structure that humans can inspect, question, and correct.

Prior work already mines or synthesizes reusable workflow artifacts. Agent Workflow Memory (AWM)~\cite{wang2024awm} induces routines from trajectories, SkillWeaver~\cite{zheng2025skillweaver} distills website practice into reusable API-style skills, AutoManual~\cite{chen2024automanual} constructs environmental manuals, ICAL~\cite{sarch2024ical} distills demonstrations into cognitive abstractions, LearnAct~\cite{liu2025learnact} studies demonstration-based mobile GUI agents, and Open-World Skill Discovery uses action-prediction error for boundary detection~\cite{deng2025openworldskill}. The broader reinforcement-learning literature gives the formal background for temporal abstraction: options and option-critic methods~\cite{sutton1999between,bacon2017option}, deep HRL systems~\cite{kulkarni2016hierarchical,vezhnevets2017feudal,nachum2018data,riemer2018learning,li2019hierarchical}, and meta-learned or offline primitives~\cite{frans2018meta,ajay2021opal}. Unsupervised skill-discovery methods such as VIC, DIAYN, DADS, CIC, and actionable representations learn reusable behaviors through mutual-information, contrastive, or goal-conditioned objectives~\cite{gregor2017vic,eysenbach2018diversity,sharma2020dads,laskin2022cic,ghosh2019actionable}; recent analysis cautions that mutual-information skills are not universally optimal for every downstream reward~\cite{eysenbach2022information}. Our negative results are consistent with this caution: a coherent skill space is not automatically a useful cross-domain policy.

Recent GUI-agent training work informs our GRPO setup. DigiRL~\cite{bai2024digirl} and WebRL~\cite{qi2025webrl} optimize agents with online RL curricula; AgentTrek~\cite{xu2025agenttrek} and OS-Genesis~\cite{sun2025osgenesis} generate agent trajectories; Proposer-Agent-Evaluator (PAE)~\cite{zhou2025pae} uses evaluator feedback for autonomous skill discovery; and Skills-Coach~\cite{tian2026skillscoach} applies a GRPO-style loop to generated task suites. Our experiment is narrower: it uses an offline IW-derived reward over text skill plans, does not interact with a live GUI during RL, and does not train the reward on target-domain task success. We therefore interpret weak GRPO transfer as a pipeline-level result, not as evidence against GUI-agent RL broadly.

\section{Problem Setup}
\label{sec:formulation}

We treat a trajectory as a sequence of UI observations and primitive actions. Primitive actions include click, type, scroll, copy, and paste. A skill label summarizes a contiguous action segment. Formally, the input dataset is
\begin{equation}
\mathcal{D} = \{\tau^{(n)}\}_{n=1}^{N}, \qquad
\tau^{(n)} = ((o_1,a_1), \ldots, (o_T,a_T)),
\end{equation}
where $o_t$ is a GUI observation and $a_t \in \mathcal{A}_{\text{low}}$ is a primitive UI action. The goal is to induce a skill vocabulary $\mathcal{Z}$ and a segmentation of each trajectory into contiguous intervals, each assigned one skill $z \in \mathcal{Z}$. In hand-authored systems, $\mathcal{Z}$ is provided by designers. In our setting, $\mathcal{Z}$ is induced from trajectories.

The question is whether the induced vocabulary $\mathcal{Z}$ helps the policy. We therefore focus the main evaluation on skill composition: whether a model can choose the right mined skill sequence on held-out tasks and transfer settings. Primitive UI-action accuracy is reported only for the verified Mind2Web zero-shot diagnostic, where benchmark annotations directly support that metric; it is not the primary Phase~3 claim.

\section{Method: Automated \texttt{SKILL.md} Generation}
\label{sec:method}

\begin{figure}[!htbp]
\centering
\includegraphics[width=0.93\textwidth]{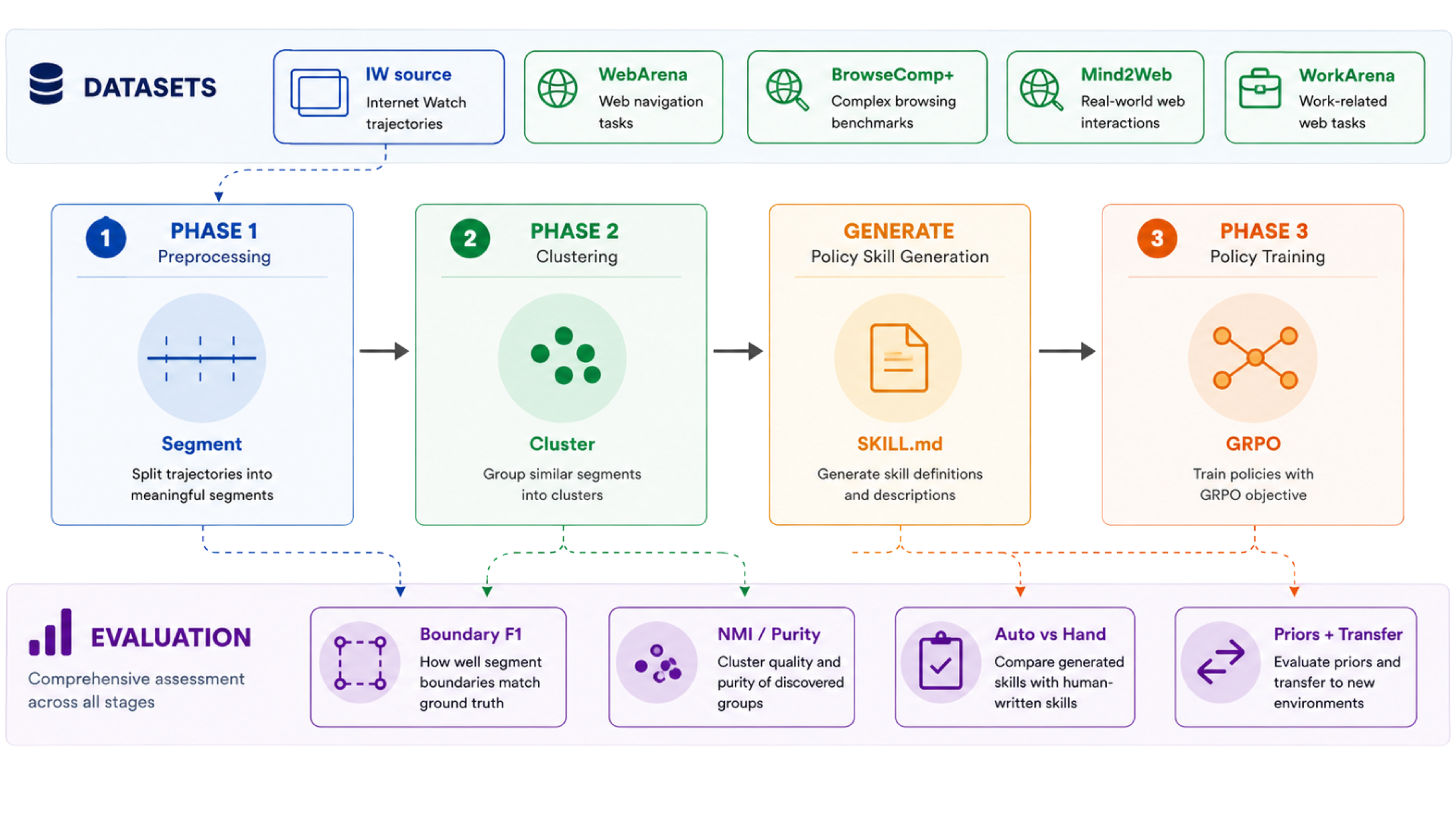}
\caption{Study design for automated \texttt{SKILL.md} generation. IW is the source dataset for trajectory segmentation, skill-library construction, and Phase~3 GRPO policy training; WebArena and BrowseComp+ are the completed held-out transfer checks. Mind2Web zero-shot and WorkArena-NLP are reported only as diagnostics, not as current GRPO transfer evidence. The paper evaluates boundary quality, cluster quality, auto-generated versus hand-crafted \texttt{SKILL.md} files, simple priors, and completed transfer checks.}
\label{fig:study_design}
\end{figure}

The pipeline has three stages. It segments trajectories, clusters the segments into skills, and trains a CUA policy with the resulting annotations. The first two stages build the skill library. The third stage tests whether the library helps. Figure~\ref{fig:study_design} summarizes the study design. The equations below are operational definitions rather than standalone theoretical claims. Equation~\ref{eq:boundary} decides where candidate skills begin and end; Equation~\ref{eq:segment-summary} turns each variable-length segment into a fixed-length vector; Equation~\ref{eq:bures} turns those vectors into a distance matrix for clustering; and Equation~\ref{eq:supcon} refines the resulting pseudo-labels into embeddings used by the sequence models. Methodologically, the pipeline combines simple offline change-point detection ideas~\cite{truong2020selective}, Gaussian optimal-transport geometry~\cite{dowson1982frechet,peyre2019computational,cuturi2013sinkhorn}, and supervised contrastive representation learning~\cite{khosla2020supervised}. The Results section organizes these tests around the main findings: boundary recall is easier than boundary precision, readable clusters remain source-bound, and the learned policies do not yet beat simple statistical priors.

\subsection{Phase 1: Trajectory Segmentation (Skill Boundary Detection)}

Given an action trajectory, we use adjacent action distance as a cheap change-point signal. For a trajectory $(a_0, a_1, \ldots, a_T)$, we compute:
\begin{equation}
\Delta a_t = \|a_t - a_{t-1}\|_2, \quad t \in \mathcal{B} \;\text{if}\; \Delta a_t > \theta,
\label{eq:boundary}
\end{equation}
where $\theta$ is selected on held-out IW data by sweeping empirical percentiles of $\Delta a_t$ and maximizing boundary F1. The boundary set $\mathcal{B}$ splits a trajectory into candidate skill segments. Each action vector has 15 normalized features: a 10-way primitive-action one-hot vector, screen coordinates $(x,y)\in[0,1]^2$, normalized timestamp, clipped text length, and clipped scroll amount. The Euclidean score is unweighted over these features and does not use DOM, screenshot, accessibility-tree, or language state. For transfer, we apply the IW-derived threshold directly; target-domain threshold sweeps are reported only as oracle diagnostics. On IW, F1 is stable over the 40th--60th percentile range (Appendix Table~\ref{tab:theta-sweep}).

The rule is intentionally simple, but it can over-split skills: click-to-type transitions, typing-length changes, and large coordinate jumps can exceed $\theta$ even when the user remains inside one skill. A natural replacement would learn boundaries from action-prediction error, as in skill-boundary detection for unsegmented demonstrations~\cite{deng2025openworldskill}; we leave that comparison for future work.

\subsection{Phase 2: Skill Embedding (Skill Library Construction)}

\paragraph{Step 1: Segment Representation.}
Let segment $\tau_i$ contain $T_i$ action vectors $a_{i,1}, \ldots, a_{i,T_i} \in \mathbb{R}^d$. We summarize the segment by the mean and diagonal variance of its action vectors:
\begin{equation}
\mu_i = \frac{1}{T_i}\sum_{t=1}^{T_i} a_{i,t}, \qquad
\Sigma_i = \mathrm{diag}\!\left(\frac{1}{T_i}\sum_{t=1}^{T_i}(a_{i,t} - \mu_i)\odot(a_{i,t} - \mu_i) + \epsilon\mathbf{1}\right),
\label{eq:segment-summary}
\end{equation}
where $\odot$ denotes element-wise multiplication and $\epsilon = 10^{-4}$ prevents degenerate variances. This is a length-invariant bag-of-actions summary. It is not a Gaussian-mixture model and it does not preserve within-segment order. This choice is useful for cheap clustering because segments with similar action inventories are close even when they have different lengths, but it discards the sequential structure that makes many GUI skills executable: selecting before copying, copying before pasting, opening a menu before choosing an item, or navigating before filling a form. Thus a cluster can be readable while still missing the order information needed for reliable downstream composition. Despite this simplification, the representation produces clusters that align with IW ground-truth skills (Section~\ref{sec:results_discovery}); the downstream sequence results test whether that bag-of-actions abstraction is sufficient for policy learning.

\paragraph{Step 2: Wasserstein Clustering.}
Let $v_i = \mathrm{diag}(\Sigma_i)$ denote the vector of diagonal variances. We group similar segments using the squared Bures distance between the diagonal Gaussians $\mathcal{N}(\mu_i, \Sigma_i)$:
\begin{equation}
D(\tau_i, \tau_j)
= \| \mu_i - \mu_j \|_2^2
+ \left\| \sqrt{v_i} - \sqrt{v_j} \right\|_2^2
= \| \mu_i - \mu_j \|_2^2
+ \sum_{k=1}^{d}\left(\sqrt{v_{i,k}}-\sqrt{v_{j,k}}\right)^2,
\label{eq:bures}
\end{equation}
This is the closed-form squared 2-Wasserstein distance for diagonal Gaussian summaries, a special case of the Fréchet/Bures distance between Gaussian measures~\cite{dowson1982frechet,peyre2019computational}. Because $v_i$ stores variances, the covariance term compares standard deviations, not raw variance vectors; using $\|v_i-v_j\|_2^2$ would be a different heuristic and is not the intended metric. The implementation follows Equation~\ref{eq:bures} by taking square roots of the diagonal variances before computing the covariance distance. We run average-linkage agglomerative clustering on this distance matrix and sweep $k$ from 8 to 16.

\paragraph{Step 3: Supervised-Contrastive Refinement.}
Let $f_\theta: \mathbb{R}^{2d} \to \mathbb{R}^{d_{\text{skill}}}$ be an MLP encoder mapping a segment summary $x_i = [\mu_i; \mathrm{diag}(\Sigma_i)]$ to an $\ell_2$-normalized embedding $z_i = f_\theta(x_i)$. Using the Wasserstein cluster assignments $c_i$ as \emph{pseudo-labels}, we train $f_\theta$ with the supervised-contrastive loss~\cite{khosla2020supervised}:
\begin{equation}
\mathcal{L}_\text{sup-con}
= \frac{1}{|\mathcal{B}|} \sum_{i\in\mathcal{B}} \; \frac{-1}{|P(i)|} \sum_{p\in P(i)}
\log \frac{\exp(z_i^\top z_p / T)}{\sum_{a\in\mathcal{B}\setminus\{i\}} \exp(z_i^\top z_a / T)},
\label{eq:supcon}
\end{equation}
where $\mathcal{B}$ is a mini-batch, $P(i) = \{p \in \mathcal{B} \setminus \{i\} : c_p = c_i\}$ is the set of same-pseudo-label positives for anchor $i$, $T = 0.07$ is a temperature, and the denominator sums over all other batch elements. We use a class-balanced sampler so each batch contains roughly equal members per pseudo-label.

The encoder is an MLP over $D_a = 15$ action-feature dimensions, so its input is the 30-dimensional vector $[\mu; \mathrm{diag}(\Sigma)]$. It maps $30 \to 64 \to 32 \to d_\text{skill}=16$ with ReLU activations and an $\ell_2$-normalized output. We train for 200 epochs with AdamW (lr=$10^{-3}$, weight decay $10^{-4}$), batch size 256, temperature $T = 0.07$, and an 80/20 stratified train/validation split by pseudo-label. The IW run takes less than five minutes on CPU.

\emph{Pseudo-supervision, not self-supervision.} Step~2 assigns labels by clustering. Step~3 trains an encoder to predict those labels through a supervised-contrastive objective. No human skill labels enter the clustering, but the encoder is still trained with pseudo-labels. This could simply overfit to the initial clusters. We test this with random labels and a k-means baseline in Section~\ref{sec:results_discovery}. Wasserstein pseudo-labels perform better than both.

\subsection{Phase 3: Skill-Aware GRPO Training}

After mining the skill library, we train models for skill-sequence composition. The lightweight baselines test whether the mined representation is useful without an LLM: an MLP receives the current skill embedding plus sequence position and predicts the next skill, while a 2-layer causal Transformer ($d_{\text{model}}=64$, 4 heads) attends over the skill sequence to test whether sequence context helps. The main policy is Qwen3-8B trained with GRPO from the base model on 1,275 prompts containing task context and mined skill names; its response format includes a skill label and a next-action field, but the reported IW/WebArena/BrowseComp+/WorkArena-NLP metrics evaluate the predicted skill sequence. A learned trajectory reward model scores the full prompt-response pair, and GRPO updates the base policy directly with no supervised warm start. We use 8 candidate responses per prompt at temperature 0.7, maximum completion length 192, learning rate $5\times10^{-6}$, gradient accumulation 8, reward clipping at 5.0, and one training epoch. The completed run took 6{,}072 seconds on 4 NVIDIA H200 NVL GPUs with 143{,}771 MiB memory per GPU.

\subsection{Transfer Evaluation}

We evaluate transfer with held-out benchmarks primarily through the IW-derived skill vocabulary. The claimed Phase~3 comparison covers IW as the source benchmark and WebArena/BrowseComp+ as held-out transfer checks. For Mind2Web, we separately report the verified zero-shot task-completion, action-accuracy, and element-accuracy diagnostics, but we do not use those numbers as evidence that the current GRPO policy improves primitive UI-action prediction. WorkArena-NLP is reported only as an auxiliary text-only planning diagnostic.

\section{Experimental Setup}
\label{sec:setup}

We evaluate the claimed pipeline on one source benchmark and two held-out transfer checks. IW is the source distribution: 2{,}000 synthetic enterprise-style trajectories with explicit skill boundaries and labels, used for segmentation, clustering, and policy training. WebArena and BrowseComp+ test whether the learned skill vocabulary transfers to external browsing traces. We also include two non-claim diagnostics: a verified Mind2Web zero-shot baseline for web-action context, and WorkArena-NLP, a text-only conversion of WorkArena task schemas into natural-language goals and structured JSON targets. Live WorkArena and current-run GRPO Mind2Web results are not part of the claimed evaluation in this submission. Appendix~\ref{app:benchmark_status} gives dataset sizes, benchmark status, and diagnostic details.

For Phase~3 skill composition, the main comparison is Qwen3-8B zero-shot versus Qwen3-8B trained with GRPO from the base model. We also include zero-shot Llama-3.1-70B, DeepSeek-R1-Distill-Qwen-32B, Gemma-4-31B, OLMo-3-7B, GPT-5, Claude Sonnet~4.5, and Claude Haiku~4.5 where available, plus simple non-LLM baselines. Segmentation is scored by boundary precision, recall, and F1; clustering by NMI, silhouette, and purity; skill composition by per-step accuracy, exact sequence match, and normalized edit distance; and Mind2Web by task completion, action accuracy, and element accuracy. Additional model, metric, and hyperparameter details are in Appendix~\ref{app:benchmark_status} and Appendix~\ref{app:impl_details}.

\section{Results}
\label{sec:results}

\subsection{Action Jumps Find Boundaries but Also Split Skills}

The first surprise is that the simplest signal works, but only in the wrong direction for a reusable skill library. Equation~\ref{eq:boundary} asks whether large adjacent action changes mark true skill switches. On IW, the best threshold is $\theta = 1.545$ (50th percentile), giving precision 0.419, recall 0.803, and F1 0.538 (Appendix Figure~\ref{fig:segmentation}). High recall means most true skill boundaries do create visible action jumps. Low precision means the same jumps also happen inside a skill: users click before typing, move between distant UI elements, scroll during review, or paste text as one step of a longer data-transfer routine. The boundary detector therefore discovers many real transitions, but it over-splits ordinary within-skill behavior.

The second part of the finding is that the threshold is not domain stable. On WebArena, applying the IW-derived threshold $\theta = 1.545$ gives precision 1.000, recall 0.100, and F1 0.119. A target-domain oracle sweep can choose $\theta = 0.603$ and reach F1 0.851, but that uses WebArena boundary labels and is not a valid zero-shot transfer result. The oracle result only shows that WebArena map-navigation trajectories contain clear discontinuities under their own action scale. It does not show that an IW-calibrated boundary rule transfers, and it should not be read as a deployable fix for Phase~1.

\subsection{The Mined Skills Are Readable but Source-Bound}
\label{sec:results_discovery}

The main positive result is narrower than expected: trajectory mining produces readable source-domain skills, but the structure does not automatically become a portable vocabulary. Equation~\ref{eq:segment-summary} compresses each discovered segment into a mean/variance action profile, and Equation~\ref{eq:bures} clusters those profiles. With the Bures metric, agglomerative clustering reaches NMI = 0.650 at $k = 8$ on IW (Appendix Table~\ref{tab:wasserstein_k}). NMI drops for larger $k$, while purity stays near 0.63, so we use $k = 8$ in the main analysis.

The pseudo-label contrastive encoder strengthens this source-domain structure. After 200 epochs, KMeans in the 16-dimensional latent space reaches NMI = 0.862, silhouette = 0.554, and purity = 0.837, a 33\% relative NMI gain over the Wasserstein baseline. Appendix Figure~\ref{fig:tsne} shows the t-SNE projection, and Appendix~\ref{app:training_curves} reports the training curve. This result shows that the cluster-derived pseudo-labels are internally consistent enough to train an embedding model.

The negative part is the transfer behavior. WebArena looks very different: with the IW threshold, segmentation F1 falls to 0.119, and the learned embedding is weak (NMI = 0.049, silhouette = $-0.255$). The target-tuned WebArena threshold can give F1 = 0.851, but that is an oracle diagnostic, not a transferable setting. The map-only trajectories do not contain enough skill diversity to form a rich library, and their discontinuity scale differs from IW. The mined taxonomy is therefore best read as a source-domain discovery result, not as evidence of a general GUI skill vocabulary.


\label{para:qualitative}
Table~\ref{tab:qualitative} explains why the result is tempting but insufficient. Five of eight clusters have purity at least 0.95 against one IW skill. Their action profiles are also interpretable: \texttt{document\_edit} contains click/format/save, \texttt{data\_transfer} contains click/copy/paste, and \texttt{organize\_files} contains click/right-click patterns. The weaker clusters, such as \texttt{send\_message} and \texttt{review\_content}, absorb broad click/type/scroll routines shared by several IW skills. The source-domain taxonomy is readable, but readability does not imply downstream usefulness. Appendix Figure~\ref{fig:qualitative} provides the full action-distribution heatmap.

\begin{table}[!htbp]
\centering
\caption{Selected $k=8$ cluster characterization on IW data. The Size column counts ground-truth segments assigned to each cluster, not primitive actions; the sizes sum to the 8{,}290 IW ground-truth segments. The action-share column lists nonzero within-cluster action percentages, and the trajectory column lists unique action-type sequences sorted by frequency. Clustering quality across different values of $k$ is moved to Appendix Table~\ref{tab:wasserstein_k}.}
\label{tab:qualitative}
\scriptsize
\setlength{\tabcolsep}{1.8pt}
\renewcommand{\arraystretch}{1.12}
\begin{tabular}{l|r|p{0.20\textwidth}|p{0.26\textwidth}|l|c}
\toprule
\textbf{Cluster} & \textbf{Size} & \textbf{Action Shares} & \textbf{Top Action Trajectories} & \textbf{Dominant GT Skill} & \textbf{Purity} \\
\midrule
C0 & 264 & click (69\%); scroll (27\%); right-click (4\%) & 1. click -> scroll -> click -> click (112); 2. click -> scroll -> click -> click -> scroll (61); 3. click -> click -> click -> click -> scroll (46) & \texttt{monitor\_status} & 0.66 \\
C1 & 2534 & click (64\%); type (36\%) & 1. click -> type -> type -> click (708); 2. click -> click -> type -> type -> click -> click (632); 3. click -> type -> click (632) & \texttt{send\_message} & 0.28 \\
C2 & 2803 & click (19\%); select (19\%); type (19\%); format (19\%); save (19\%); scroll (6\%) & 1. click -> select -> type -> format -> save (1963); 2. click -> select -> type -> format -> save -> scroll (840) & \texttt{document\_edit} & \textbf{1.00} \\
C3 & 110 & click (38\%); copy (19\%); switch (19\%); paste (19\%); scroll (6\%) & 1. click -> copy -> switch -> click -> paste (74); 2. click -> copy -> switch -> click -> paste -> scroll (36) & \texttt{data\_transfer} & \textbf{1.00} \\
C4 & 111 & click (75\%); right-click (25\%) & 1. click -> right-click -> click -> click (111) & \texttt{organize\_files} & \textbf{1.00} \\
C5 & 71 & click (100\%) & 1. click -> click -> click -> click (71) & \texttt{export\_publish} & \textbf{1.00} \\
C6 & 2176 & click (54\%); type (24\%); scroll (22\%); save (1\%) & 1. click -> scroll -> click -> type -> click (694); 2. click -> scroll -> click -> type -> click -> scroll (312); 3. click -> type -> click -> scroll (300) & \texttt{review\_content} & 0.46 \\
C7 & 221 & click (60\%); type (20\%); save (20\%) & 1. click -> click -> type -> click -> save (221) & \texttt{presentation\_edit} & \textbf{1.00} \\
\bottomrule
\end{tabular}
\end{table}

\subsection{Learned Skill Composition Does Not Beat Simple Priors}
\label{sec:results_composition}

The strongest negative result appears after the readable clusters are used for prediction. The MLP and Transformer consume the learned segment embeddings $z_i$, while the GRPO policy consumes text prompts built from the same mined skill labels. Appendix Table~\ref{tab:composition} compares these pipeline variants on IW, but it is not a controlled architecture-only or reward-model ablation: input modality, model class, supervision signal, and optimizer all change at once. The comparison is still useful as a sanity check, because every variant is supposed to benefit from the mined skill structure.

The sanity check fails for the main learned policy. The GRPO-trained Qwen3-8B policy reaches 20.5\% skill-step accuracy, below the MLP baseline (23.3\%), the Transformer baseline (34.6\%), and the trivial Frequency baseline in Appendix Table~\ref{tab:baselines} (34.9\%). Exact sequence match is 0\% for the GRPO policy. The result should be interpreted as a failure of the current text-prompted GRPO setup as a whole, not as proof that the reward model alone is responsible. A reward-specific conclusion would require a modality-matched control, such as Qwen3-8B trained on the same text prompts with supervised next-skill loss, or an embedding-based policy trained with and without the same reward model.

One likely reason is that the useful part of a GUI skill is often ordered, while the segment embedding is not. The Phase~2 representation summarizes a segment by mean and variance, so it does not record whether a click happened before or after a type, copy, paste, or menu action. Per-skill accuracy follows this pattern: it ranges from 94.1\% for \texttt{search\_navigate} to 55.6\% for \texttt{presentation\_edit} (Table~\ref{tab:perskill}). Skills with distinct action signatures are easier to predict; skills that share click/type/scroll routines are harder. Appendix Figure~\ref{fig:composition} shows the per-position MLP and Transformer comparison.

\subsection{GRPO Does Not Establish Transfer on Completed Checks}

Table~\ref{tab:sixmodel} gives the completed benchmark-level view. IW is the source benchmark; WebArena and BrowseComp+ are the held-out transfer checks used for the paper's policy-transfer claim. WorkArena-NLP is included in the same table only as an auxiliary text-only planning diagnostic, not as a substitute for live WorkArena. The API models are zero-shot format-constrained baselines. The current GRPO run improves IW skill-step accuracy only slightly over zero-shot Qwen3-8B (18.5\% to 20.5\%), decreases on WebArena (55.8\% to 44.2\%), is unchanged on BrowseComp+ (43.5\% to 43.3\%), and matches zero-shot Qwen3-8B on WorkArena-NLP field accuracy (37.0\% for both, with 0\% exact match). We therefore limit the transfer conclusion to the completed WebArena and BrowseComp+ checks.

This is surprising because the training signal is built around the mined skills, but the verified transfer gains are absent or negative. Closed-source models improve the prompting baseline but do not dominate every benchmark: GPT-5 and Claude Sonnet~4.5 are strongest on WebArena, while OLMo-3-7B remains strongest on BrowseComp+. WorkArena-NLP completed runs range from OLMo-3-7B at 12.8\% to GPT-5 at 40.6\%, with DeepSeek-R1-Distill-32B at 31.2\% and both Qwen3-8B variants at 37.0\%; these numbers diagnose text-only schema recovery, not live enterprise automation. Appendix Table~\ref{tab:baselines} reinforces the negative result: a trivial Frequency baseline is strong on IW, strictly outperforming both the MLP and GRPO policies on skill-step accuracy. The current learned skill-composition methods have therefore not shown value beyond dataset class imbalance.

\begin{table}[!htbp]
\centering
\caption{Completed model comparison for composing auto-generated skills. IW is the source benchmark; WebArena and BrowseComp+ are the held-out transfer checks used for the policy-transfer claim. Scores for these three columns are single-run skill-step percentages (\%). WorkArena-NLP is an auxiliary text-only diagnostic and reports field accuracy on 100 structured-planning examples; it is not a live WorkArena substitute. API models are zero-shot format-constrained prompting baselines. The DeepSeek BrowseComp+ score is a verified near-zero result: 1/2{,}194 correct samples, or 0.0456\%, rounded to 0.05. Best per-column scores are shown in \textbf{bold}.}
\label{tab:sixmodel}
\small
\begin{tabular}{lcccc}
\toprule
\textbf{Model} & \textbf{IW} & \textbf{WebArena} & \textbf{BrowseComp+} & \textbf{WorkArena-NLP} \\
\midrule
Qwen3-8B (zero-shot) & 18.5 & 55.8 & 43.5 & 37.0 \\
Qwen3-8B (GRPO from base) & 20.5 & 44.2 & 43.3 & 37.0 \\
Llama-3.1-70B & \textbf{30.0} & 56.2 & 51.9 & 38.0 \\
OLMo-3-7B & 14.3 & 54.5 & \textbf{61.4} & 12.8 \\
Gemma-4-31B & 16.5 & 37.7 & 11.2 & 38.6 \\
DeepSeek-R1-Distill-32B & 14.3 & 45.1 & 0.05 & 31.2 \\
GPT-5 & 24.5 & \textbf{57.6} & 59.5 & \textbf{40.6} \\
Claude Sonnet~4.5 & 25.8 & \textbf{57.6} & 47.7 & 39.8 \\
Claude Haiku~4.5 & 27.8 & 50.8 & 51.0 & 39.6 \\
\bottomrule
\end{tabular}
\end{table}

Mind2Web is not included in Table~\ref{tab:sixmodel} because the current GRPO run does not have a completed Mind2Web evaluation. Appendix~\ref{sec:results_e2e} reports only a zero-shot baseline for context.

\subsection{Auto-\texttt{SKILL.md} Can Beat a Manual Table but Not Frequency}
\label{sec:results_autoskillmd}

\begin{figure}[!htbp]
\centering
\includegraphics[width=\textwidth]{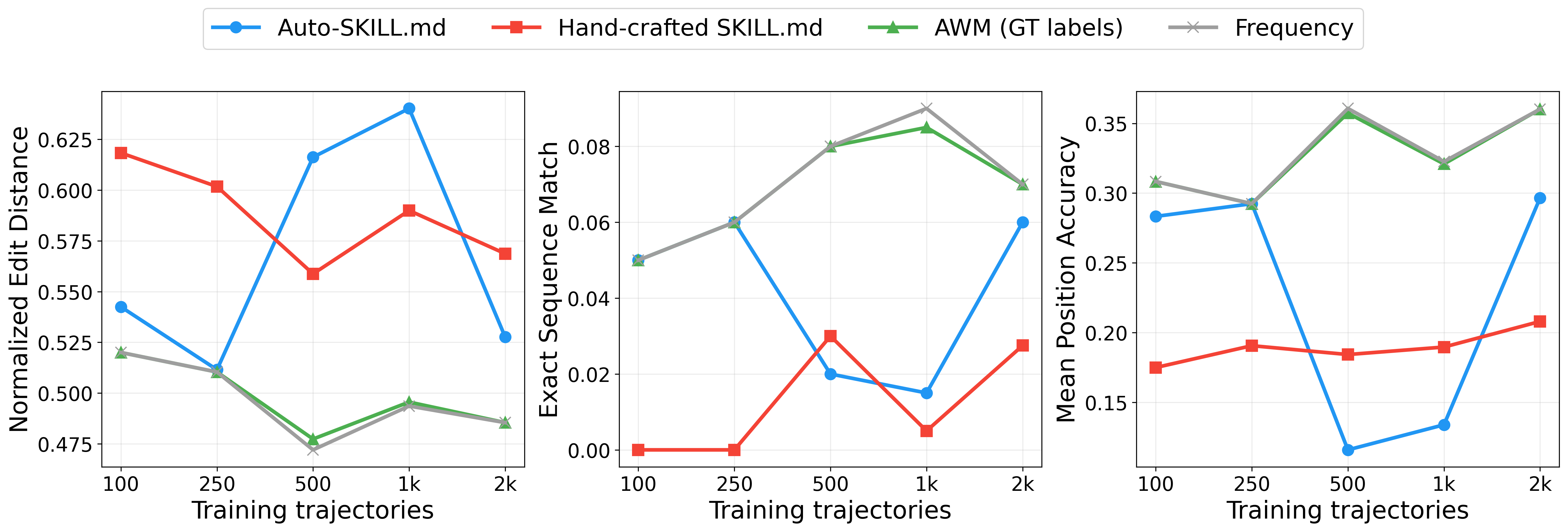}
\caption{Data-efficiency comparison for generated \texttt{SKILL.md}. Lower normalized edit distance is better; higher exact match and mean position accuracy are better. The generated specification improves over the hand-written baseline at some sizes, including the largest setting, but it remains worse than the Frequency baseline across the evaluated sizes. This supports a cautious conclusion: trajectory-mined specifications can be competitive with simple manual tables, but the current pipeline is not yet a reliable replacement for trivial statistical baselines or manual design.}
\label{fig:autoskillmd_efficiency}
\end{figure}

The final result concerns the artifact that motivated the pipeline: generated \texttt{SKILL.md} files. We run the system on IW subsets and generate skill descriptions, transition probabilities, workflows, and error-handling patterns, then compare these files with hand-written baselines. Appendix~\ref{app:skillmd_examples} gives three same-task qualitative comparisons.

Figure~\ref{fig:autoskillmd_efficiency} and Appendix Table~\ref{tab:autoskillmd} compare generated and hand-written \texttt{SKILL.md} files as the number of IW training trajectories changes. The important question is not whether the generated transition table looks different from the hand-written one, but whether it improves next-skill prediction. Relative to the hand-written table, the result is mixed: Auto-generated \texttt{SKILL.md} is better at $N = 100$, $250$, and $2{,}000$, but worse at $N = 500$ and $1{,}000$.

The negative finding is sharper when the trivial baseline is included. Frequency has lower normalized edit distance at every evaluated training size; at $N = 2{,}000$, Frequency reaches 0.485 while Auto-\texttt{SKILL.md} reaches 0.528. Thus automated generation can beat a simple hand-written table in some settings, but it does not yet beat the strongest trivial baseline. We therefore report the data-efficiency curve in Figure~\ref{fig:autoskillmd_efficiency}, which compares methods on normalized edit distance, exact sequence match, and mean position accuracy.



\section{Discussion}
\label{sec:discussion}


Our results demonstrate that mined-skill methods should be evaluated against frequency and transition priors before being compared to large language models. In our experiments, the most-common-skill baseline is not a strawman: it captures class imbalance and repetitive workflow structure that learned systems can easily appear to exploit. Future work on GUI skill discovery should therefore report at least three controls: a frequency prior, a transition-memory prior, and a modality-matched supervised policy. Without these controls, improvements from mined skill labels may be mistaken for improvements from dataset imbalance or output-format adaptation.


The small Transformer beats the GRPO-trained Qwen3-8B policy on IW, but that comparison changes input modality and optimization objective at the same time: the Transformer consumes continuous segment embeddings and is trained directly for sequence prediction, while Qwen3-8B consumes text trajectory prompts and is optimized through a learned trajectory reward model. These results show that the current pipeline can produce readable transition priors, but they do not show that the learned components capture reusable skill-composition structure beyond dataset imbalance.

The limitations point to concrete next steps. The contrastive encoder uses cluster-derived pseudo-labels, so the pipeline is not fully unsupervised; IW is synthetic and may not capture real enterprise complexity; and the Phase~1 $\ell_2$ boundary heuristic should be compared against learned action-prediction-error segmenters before being treated as robust. Stronger claims would require completed Mind2Web GRPO and live WorkArena evaluations; the present submission does not make those claims. Better follow-up experiments would mix IW with target-domain prompts, replace the bag-of-actions segment representation with an order-aware encoder, train a denser action-level or task-success reward model, and run factorial ablations that hold two stages fixed while replacing the third with ground-truth boundaries, oracle target-domain boundaries, or supervised controls. Because CUAs can automate useful repetitive work but can also be misused, generated \texttt{SKILL.md} artifacts should remain reviewable by humans before deployment.

\section{Conclusion}

We mined skill specifications from GUI trajectories and tested what transfers. The mined skills are readable on the source data: five of eight clusters reach at least 0.95 purity. The trained policy does not transfer well. GRPO from base Qwen3-8B reaches 20.5\% IW skill-step accuracy versus 18.5\% zero-shot, and 43.3\% BrowseComp+ skill-step accuracy versus 43.5\% zero-shot. The learned components also fail important sanity checks: a Frequency baseline beats the proposed MLP and GRPO policies on IW skill-step accuracy, and it beats Auto-\texttt{SKILL.md} on edit distance across all evaluated data sizes. Mined skills are useful as inspectable structure and reward-model training signal. They are not yet a reliable replacement for trivial statistical baselines, manual design, or a cross-domain policy.

\bibliographystyle{unsrtnat}
\bibliography{citation}

\newpage
\appendix
\setcounter{table}{0}
\renewcommand{\thetable}{A\arabic{table}}
\setcounter{figure}{0}
\renewcommand{\thefigure}{A\arabic{figure}}

\section{Appendix}
\label{app:impl_details}


\subsection{Additional GRPO Training Sessions}

We also run a scale-control GRPO session on Llama-3.1-70B-Instruct with quantized low-rank adaptation (QLoRA) adapters. This anti-collapse rerun starts from the base model, uses 2 candidate responses per prompt, maximum completion length 64, learning rate $10^{-5}$, gradient accumulation 16, and one epoch. The reward is a hand-weighted skill-format reward with components \texttt{correct\_skill(+1.0)}, \texttt{format(+0.1)}, \texttt{reasoning(+0.1)}, and \texttt{invalid(-0.3)}. The run completes 1{,}149 optimizer steps in 224{,}341 seconds on 8 NVIDIA RTX A6000 GPUs with 49{,}140 MiB memory per GPU, with final training loss $-1.16\times10^{-4}$, mean reward 0.529, reward standard deviation 0.460, and mean entropy 0.277. 

\subsection{Hyperparameters}
\label{app:hyperparams}

Table~\ref{tab:hyperparams} summarizes the model and optimizer settings used for the encoder, sequence baseline, and GRPO runs.

\begin{table}[h!]
\centering
\caption{Complete hyperparameters for all model configurations.}
\label{tab:hyperparams}
\small
\begin{tabular}{lll}
\toprule
\textbf{Component} & \textbf{Hyperparameter} & \textbf{Value} \\
\midrule
\multirow{3}{*}{Phase 2 Encoder} & Architecture & MLP: $30 \to 64 \to 32 \to 16$ \\
& Optimizer & AdamW, lr=$10^{-3}$, wd=$10^{-4}$ \\
& Training & 200 epochs, batch 256, $T=0.07$ \\
\midrule
\multirow{3}{*}{Phase 3 Transformer} & Architecture & 2-layer, 4 heads, $d=64$ \\
& Optimizer & Adam, lr=$10^{-4}$ \\
& Training & 500 epochs, teacher forcing \\
\midrule
\multirow{12}{*}{GRPO} & Base model & Qwen3-8B \\
& Candidates per prompt & 8 \\
& Temperature & 0.7 \\
& Optimizer & cosine, lr=$5\times10^{-6}$ \\
& Training & 1 epoch, 4 GPUs, 6{,}072 seconds \\
& Reward & trajectory reward model, clip=5.0 \\
\cmidrule{2-3}
& Base model & Llama-3.1-70B-Instruct \\
& Candidates per prompt & 2 \\
& Maximum completion length & 64 \\
& Optimizer & cosine, lr=$10^{-5}$ \\
& Training & 1 epoch, 8 GPUs, 224{,}341 seconds \\
& Reward & hand-weighted skill-format reward \\
\bottomrule
\end{tabular}
\end{table}

\subsection{Assets, Licenses, and New Artifacts}
\label{app:assets}

We use third-party benchmarks and models only as cited research artifacts or through their public provider interfaces. The new dataset introduced by this work are the synthetic IW trajectories, mined cluster assignments, generated \texttt{SKILL.md} specifications, evaluation prompts/predictions derived from our runs, and the WorkArena-NLP diagnostic conversion. 

Section~\ref{sec:setup} documents the dataset composition and curation logic; Appendix~\ref{app:benchmark_status} documents benchmark status and diagnostics; Appendix~\ref{app:generated_skill_specs} documents generated skill specifications and examples. Code and public-release artifacts are available through the anonymous project repository linked in the main text.

\subsection{GRPO Reward Model}
\label{app:grpo_reward}

The current GRPO run uses a learned trajectory reward model rather than a hand-weighted format reward. The reward model is a Qwen3-8B transformer with a scalar sequence-classification head, loaded through \texttt{AutoModelForSequenceClassification(num\_labels=1)}. For an input string $x$, it returns one real-valued logit $r_\phi(x)$. It is trained with the Transformer Reinforcement Learning (TRL) library's \texttt{RewardTrainer} on pairwise preferences over full skill plans. The input text is the same planning prompt used by the policy concatenated with a candidate completion of the form \texttt{[Plan] skill\_1 -> skill\_2 -> ...}.

The reward-model data comes from the train and validation JSONL files under \texttt{data/interaskill/conversations/}. These files contain 1{,}275 train conversations and 225 validation conversations with task text, message history, and a ground-truth \texttt{skill\_flow}. We keep examples with at least two skills. For each prompt, the ground-truth skill flow is the chosen completion. Rejected completions are synthetic hard negatives generated by adjacent swaps, single-skill replacements, deletions, insertions, global swaps, and duplicate/mutation operations. The training run uses three negatives per train prompt and two negatives per validation prompt, producing approximately 3{,}825 train preference pairs and 450 validation pairs.

Preference margins are generated heuristically rather than from held-out benchmark correctness. A candidate plan is scored against the ground-truth skill flow by a weighted combination of prefix match (0.45), longest-common-subsequence overlap (0.30), unordered skill overlap (0.20), and length agreement (0.05). The chosen plan receives the self-score; each rejected plan receives its heuristic score, and the preference margin $m$ is clipped to $[0.05, 0.95]$. TRL's reward training objective is a pairwise logistic ranking loss over the chosen and rejected scalar logits, with the margin requiring the chosen score to exceed the rejected score by more than the heuristic gap:
\begin{equation}
\mathcal{L}_{\mathrm{RM}}
= -\log \sigma\!\left(r_\phi(p, y^+) - r_\phi(p, y^-) - m\right),
\end{equation}
where $p$ is the prompt, $y^+$ is the ground-truth skill-flow completion, and $y^-$ is a synthetic near-miss plan. Thus the reward model learns to rank IW-style skill plans by similarity to the annotated skill flow. It is not trained on WebArena, WorkArena, BrowseComp+, Mind2Web, or live task-success labels.

There is no separate methodology for labeling ``held-out benchmark correctness'' for reward-model training: those labels are not used. Validation during reward-model training uses held-out IW conversations from the validation split in \texttt{data/interaskill/conversations/} with the same synthetic preference construction as training. Held-out benchmark correctness is measured only after GRPO by running the policy evaluator on the completed IW, WebArena, and BrowseComp+ checks. This distinction is important because the learned reward is a proxy for IW skill-flow similarity, not an estimator of cross-domain task success.

Reward-model training uses Qwen/Qwen3-8B, maximum length 768, one epoch, per-device batch size 1, gradient accumulation 32 in the reported script, learning rate $10^{-5}$, weight decay 0.01, warmup ratio 0.1, cosine scheduling, bf16, and gradient checkpointing. The reported script can optionally load the reward backbone in 4-bit, but the final trajectory reward model used by GRPO is saved as a standard sequence-classification checkpoint. During GRPO, the reward function concatenates the policy prompt and completion, tokenizes to maximum length 1024, runs the reward model, and returns the scalar logit. The reported GRPO run clips this scalar to $[-5, 5]$ before within-group advantages are computed. This design avoids directly rewarding superficial format features, but it means reward quality is limited by the IW skill-flow preference construction and by the mismatch between text-plan rewards and downstream benchmark correctness.

\section{Benchmark Status and Transfer Diagnostics}
\label{app:benchmark_status}

\subsection{Detailed Experimental Setup}

\paragraph{Datasets.} The claimed pipeline evaluation uses one source dataset and two held-out transfer checks, with two additional diagnostics. IW is the source dataset because it provides ground-truth skill boundaries and labels: 2{,}000 fabricated interaction trajectories spanning 12 skill types (e.g., \texttt{document\_edit}, \texttt{search\_navigate}, \texttt{send\_message}), totaling 8{,}290 ground-truth segments and 40{,}774 primitive actions. We curate IW to cover repeated enterprise routines such as document editing, communication, file organization, status monitoring, and cross-application data transfer. WebArena~\cite{zhou2024webarena} contributes 1{,}000 real map-navigation trajectories with 3{,}140 segments; it is useful as a stress test for realistic but homogeneous web behavior. BrowseComp+ contains trajectories derived from BrowseComp and tests complex browsing transfer with the same skill-aware response format as IW. Mind2Web~\cite{deng2023mind2web} is used only for a verified zero-shot diagnostic in this submission. WorkArena~\cite{drouin2024workarena} is the closest enterprise transfer benchmark, but live WorkArena results are not claimed; WorkArena-NLP is a text-only diagnostic constructed by converting WorkArena task schemas into natural-language goals paired with structured JSON targets across navigation, sorting, filtering, record creation, and service-catalog ordering.

\paragraph{Models and baselines.} For Phase~3 skill composition, we evaluate Qwen3-8B in zero-shot and GRPO-trained settings. We compare against zero-shot Llama-3.1-70B, DeepSeek-R1-Distill-Qwen-32B, Gemma-4-31B, OLMo-3-7B, GPT-5, Claude Sonnet~4.5, and Claude Haiku~4.5. We additionally include three non-LLM baselines: a manually authored \texttt{SKILL.md} transition table, AWM-style bigram transitions~\cite{wang2024awm}, and a most-frequent-skill predictor.

\paragraph{Metrics.} For Phase~1 segmentation we report boundary precision, recall, and F1 against ground-truth skill boundaries. For Phase~2 clustering we report Normalized Mutual Information (NMI), silhouette score, and purity, both on the raw Wasserstein clusters and on the supervised-contrastive refined latent space. For Phase~3 skill composition we report per-position accuracy, exact sequence match rate, and normalized edit distance between predicted and ground-truth skill sequences. Per-position accuracy reveals error compounding along the sequence; edit distance captures the cost of fixing a predicted sequence in tokens. For the Mind2Web zero-shot diagnostic, a task is counted as complete under task completion rate (TCR) only when every predicted action matches the ground-truth action in sequence; we additionally report per-step action accuracy and element accuracy.

\subsection{Additional Transfer Tables}
\label{app:additional_tables}

Tables~\ref{tab:baselines}--\ref{tab:e2e_domain} provide supporting transfer results that are useful for context but are not the headline claims in the main text. They also expose a negative result: on IW, the most-common-skill Frequency baseline is stronger than the proposed MLP and GRPO policies, so these learned methods should not be interpreted as solving skill composition.

\begin{table}[t]
\centering
\caption{Auto-generated vs.\ hand-crafted skill specifications. IW Acc. is InteraSkill Workflows skill-sequence accuracy and WebArena (WA) Acc. is WebArena skill-sequence accuracy. Mean position accuracy is reported. The hand-crafted baseline is a simple transition table, not an optimized expert system. The Frequency baseline is trivial but strong, outperforming the proposed MLP and GRPO policies on IW skill-step accuracy.}
\label{tab:baselines}
\small
\begin{tabular}{llccc}
\toprule
\textbf{Model} & \textbf{Type} & \textbf{IW Acc.} & \textbf{WA Acc.} & \textbf{IW Edit Dist.} \\
\midrule
SKILL.md & Fixed table & 0.140 & 0.087 & 0.633 \\
Frequency & Most common & 0.349 & 0.285 & 0.480 \\
AWM~\cite{wang2024awm} & Learned transitions & 0.334 & \textbf{0.788} & 0.479 \\
\midrule
MLP (ours) & Embedding-based & 0.233 & 0.285 & 0.557 \\
Transformer (ours) & Sequence model & 0.346 & 0.410 & 0.481 \\
Qwen3-8B GRPO (ours) & Policy optimization & 0.205 & 0.442 & 0.672 \\
\bottomrule
\end{tabular}
\end{table}

\begin{table}[t]
\centering
\caption{Skill composition on IW data using mined skills. This table is not a controlled input-modality or reward-model ablation: the MLP and Transformer operate on continuous segment embeddings, while the GRPO-trained Qwen3-8B policy operates on text trajectory prompts and is optimized through a learned reward model. The rows should therefore be read as pipeline variants, not as evidence that the reward model alone causes the performance gap.}
\label{tab:composition}
\small
\begin{tabular}{lccccc}
\toprule
\textbf{Model} & \textbf{Accuracy} & \textbf{Exact Match} & \textbf{Edit Dist.} & \textbf{Params} & \textbf{Training Data} \\
\midrule
MLP & 0.233 & 0.054 & 0.557 & 5.6K & Embeddings \\
Transformer & 0.346 & 0.079 & 0.481 & 45K & Embeddings \\
Qwen3-8B GRPO & 0.205 & 0.000 & 0.672 & 8B & Trajectory prompts \\
\bottomrule
\end{tabular}
\end{table}

\begin{figure}[t]
\centering
\includegraphics[width=0.85\textwidth]{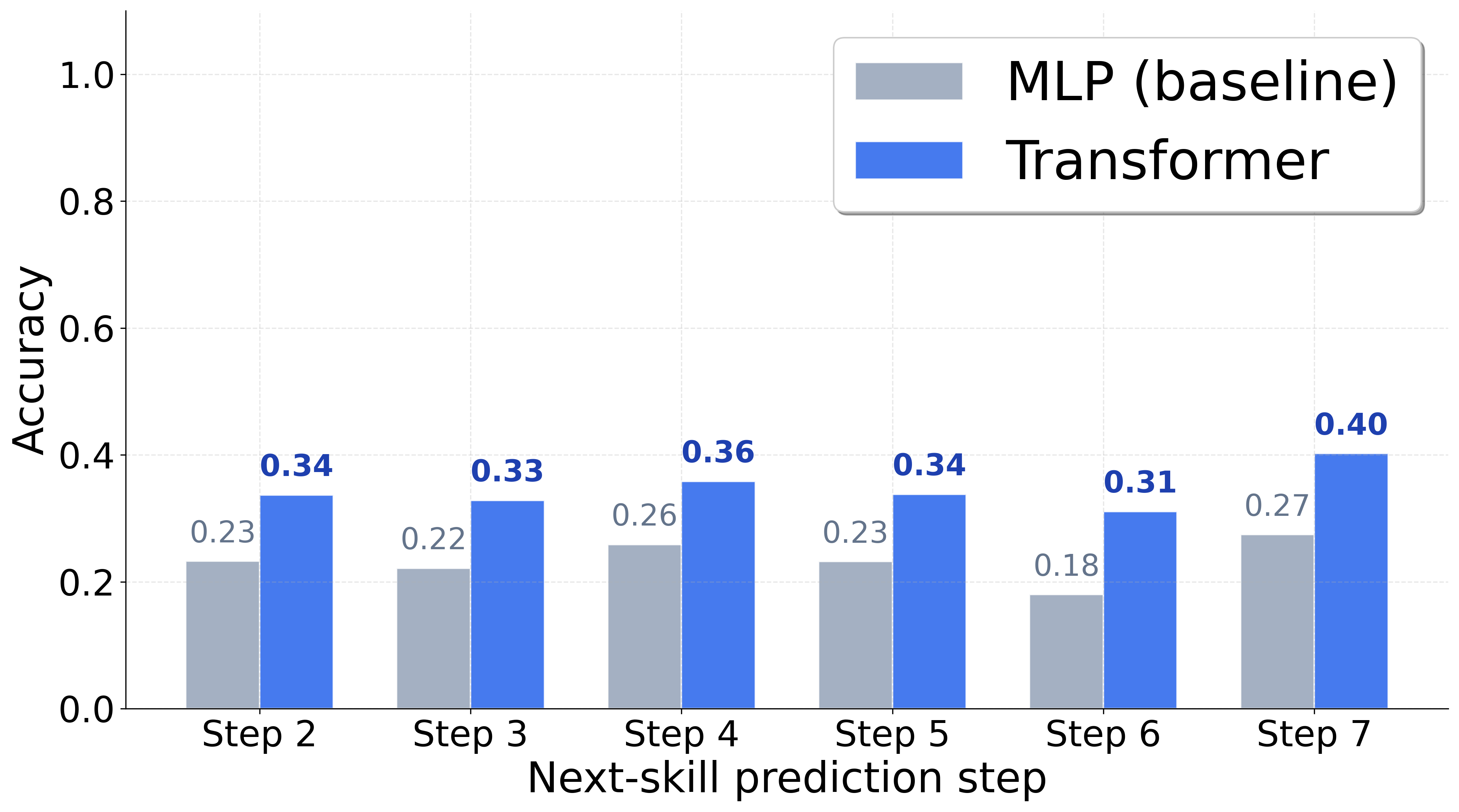}
\caption{Next-skill prediction accuracy for MLP versus Transformer on IW data. Step~2 predicts the second skill given the ground-truth first skill; the seeded first skill at Step~1 is omitted. Both models operate on learned skill embeddings from Phase~2. Accuracy degrades at later steps due to error compounding, with the Transformer's sequential attention providing consistent improvement.}
\label{fig:composition}
\end{figure}

\begin{table}[t]
\centering
\caption{Verified Mind2Web zero-shot baseline. The current GRPO run requires a fresh Mind2Web evaluation before it can be compared here.}
\label{tab:e2e}
\small
\begin{tabular}{lcccc}
\toprule
\textbf{Configuration} & \textbf{TCR} & \textbf{Reward} & \textbf{Action Acc.} & \textbf{Element Acc.} \\
\midrule
Qwen3-8B zero-shot (test\_task) & 9.5\% & 0.273 & 28.8\% & 33.8\% \\
\bottomrule
\end{tabular}
\end{table}

\begin{table}[t]
\centering
\caption{Per-domain Mind2Web zero-shot results (test\_task split).}
\label{tab:e2e_domain}
\small
\begin{tabular}{lcc}
\toprule
\textbf{Domain} & \textbf{$n$} & \textbf{TCR (zero-shot)} \\
\midrule
Entertainment & 57 & 8.8\% \\
Shopping & 63 & 19.0\% \\
Travel & 80 & 2.5\% \\
\bottomrule
\end{tabular}
\end{table}

\subsection{WorkArena Diagnostic Status}
\label{app:workarena_status}

Live WorkArena is not part of the claimed evaluation in this submission. It would be a stronger test of enterprise transfer than WebArena because its tasks involve knowledge-base lookup, record editing, forms, lists, dashboards, and service catalog workflows, which are close to the workflow types that IW tries to simulate.

We added an offline WorkArena evaluation path for future use. It uses the same conversation format as IW, WebArena, and BrowseComp+. It expects WorkArena rollouts under \texttt{data/workarena/trajectories/}, converts them to JSONL conversations under \texttt{data/workarena/conversations/}, and evaluates skill prediction with \texttt{interaskill.eval\_model}. Because the rollout file has not been collected, no live WorkArena result is reported or used in the paper's transfer claims.

We also construct a text-only WorkArena-NLP variant for diagnosing whether models understand the enterprise task specifications independently of browser control. On the first 100 examples, completed runs reach 12.8--40.6\% field accuracy and 0\% exact match, showing that partial schema recovery is possible but full structured reconstruction remains difficult. This is not a replacement for live WorkArena: it removes visual grounding, state tracking, clicking, and form interaction. It instead tests instruction understanding and workflow planning. Appendix~\ref{app:workarena_nlp} gives the conversion details.

\subsection{Text-Only WorkArena Conversion}
\label{app:workarena_nlp}

WorkArena is originally a live computer-using-agent benchmark. A model observes a ServiceNow page and must issue browser actions such as \texttt{click}, \texttt{fill}, \texttt{select\_option}, \texttt{scroll}, and \texttt{press}. The ground truth is therefore an environment state, not a text label. For example, the basic menu-navigation task is correct only when the browser reaches the exact target ServiceNow URL; form and list tasks are correct only when the corresponding record, filter, sort order, or catalog order exists in the web application.

To separate task understanding from browser grounding, we define a derived text-only benchmark, WorkArena-NLP. We do not use screenshots, accessibility trees, or live browser state. Instead, we read WorkArena's configuration files and task schemas and convert each task into a pair $(g, y)$, where $g$ is the natural-language goal shown to the agent and $y$ is a structured JSON plan containing the fields that the live task evaluator would eventually check. The model is prompted with $g$ and must output $y$.

The conversion preserves the semantic target of each task family:
\begin{itemize}\setlength\itemsep{0.2em}
    \item Navigation tasks from \texttt{all\_menu.json} become structured plans with \texttt{task\_type}, \texttt{application}, \texttt{module}, \texttt{target\_url}, and \texttt{required\_final\_action}.
    \item Sorting and filtering tasks expose the list name, filter kind, filter fields, filter values, sort fields, and sort directions.
    \item Record-creation tasks expose the ServiceNow table, requested fields, field labels, target values, and \texttt{required\_final\_action=submit\_record}.
    \item Service-catalog orders expose the item, description, quantity, item configuration values, and \texttt{required\_final\_action=order\_item}.
\end{itemize}

The generated dataset contains 2{,}210 examples: 1{,}000 navigation examples, 300 sort examples, 300 filter examples, 250 create-record examples, and 360 catalog-order examples. We score model outputs by parsing the returned JSON and comparing it with the expected plan. Exact match requires all flattened fields to match after simple string normalization. Field accuracy is the fraction of expected flattened fields that match. This metric is intentionally stricter than semantic similarity but easier to audit.

WorkArena-NLP should be interpreted as a diagnostic benchmark. It answers whether a model can parse an enterprise workflow instruction into the right structured intent. It does not evaluate visual grounding, dynamic state tracking, login/session handling, menu expansion, clicking, form entry, or recovery from web-interface errors. Thus, WorkArena-NLP results are complementary to live WorkArena results rather than substitutes for them.

\subsection{Mind2Web Diagnostic Status}
\label{sec:results_e2e}

Mind2Web would be a stronger test of cross-domain web action transfer than the skill-sequence checks in Table~\ref{tab:sixmodel}, but the current GRPO run does not have a completed Mind2Web evaluation. Table~\ref{tab:e2e} therefore reports only the verified Mind2Web zero-shot baseline as context, not as evidence for the proposed policy.

Base Qwen3-8B reaches 9.5\% task completion, 0.273 reward, 28.8\% action accuracy, and 33.8\% element accuracy on Mind2Web test\_task. A future GRPO evaluation would need to beat this number before the paper could claim Mind2Web transfer. Zero-shot performance varies by domain: Shopping is easiest and Travel is hardest (Table~\ref{tab:e2e_domain}). The present submission makes no Mind2Web or live WorkArena transfer claim.

\label{para:skill-action-gap}
The current result does not demonstrate whether the IW skill vocabulary transfers to Mind2Web. It only shows that the source-domain clusters are readable and that the current policy result is weak on the benchmarks we verified.



\section{Skill Discovery Diagnostics}
\label{app:skill_diagnostics}
\subsection{Segmentation Threshold Sensitivity}

Table~\ref{tab:theta-sweep} reports the source-domain sweep used to select the action-discontinuity threshold $\theta$.

\begin{table}[h!]
\centering
\caption{Sensitivity of segmentation F1 to the percentile used for $\theta$ on IW. Stable within a $\pm10$-percentile window around the optimum.}
\label{tab:theta-sweep}
\small
\begin{tabular}{ccc}
\toprule
Percentile & $\theta$ & F1 \\
\midrule
40 & 1.41 & 0.527 \\
45 & 1.49 & 0.534 \\
50 & 1.545 & \textbf{0.538} \\
55 & 1.60 & 0.535 \\
60 & 1.66 & 0.524 \\
\bottomrule
\end{tabular}
\end{table}

\begin{figure}[t]
\centering
\includegraphics[width=0.72\textwidth]{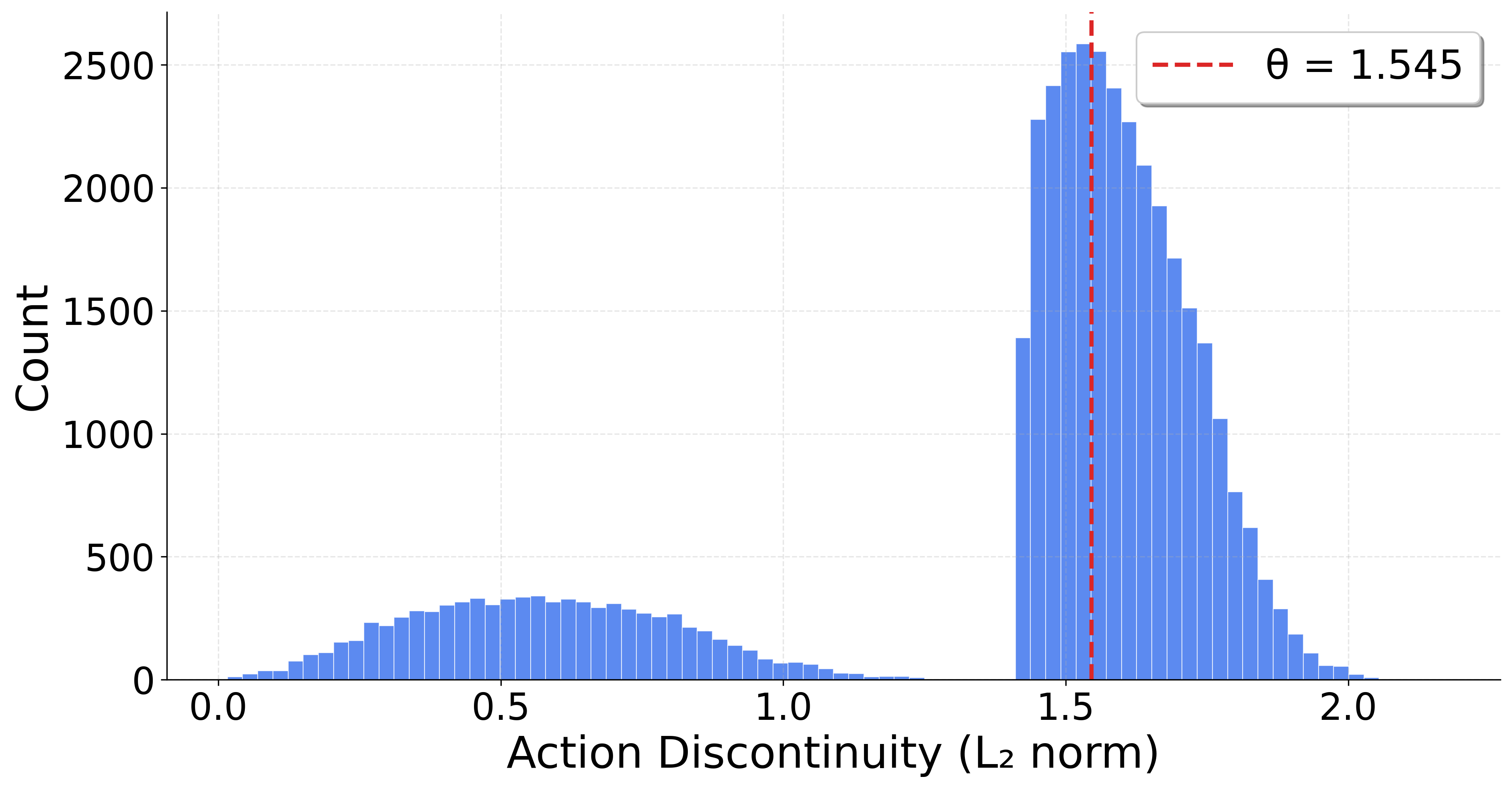}
\caption{IW action-discontinuity scores. The dashed line marks the selected source-domain threshold $\theta = 1.545$.}
\label{fig:segmentation}
\end{figure}

\subsection{Embedding Visualization}

\begin{figure}[h!]
\centering
\includegraphics[width=0.82\textwidth]{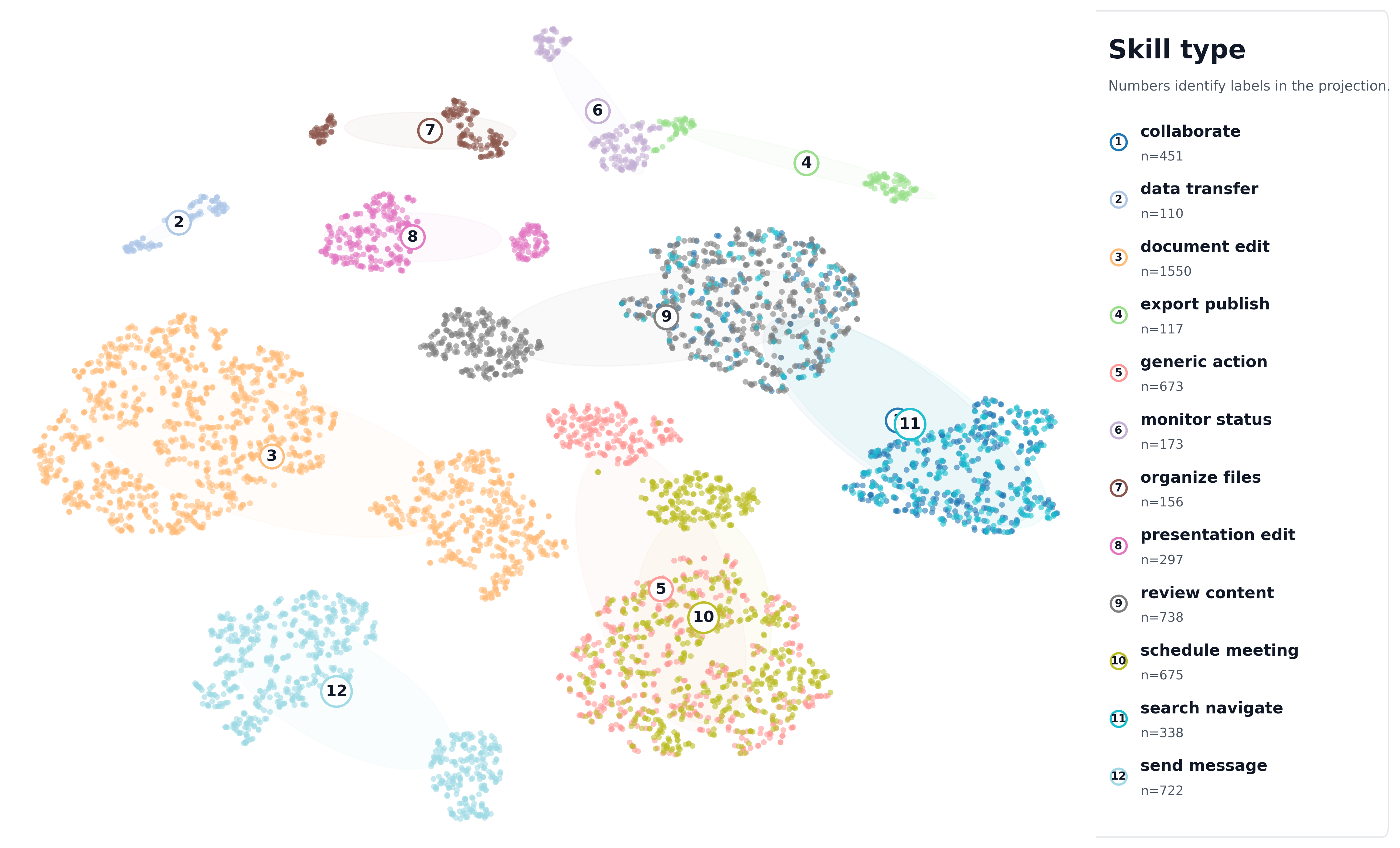}
\caption{t-distributed stochastic neighbor embedding (t-SNE) projection of 16-dim skill embeddings on the IW benchmark. Colors and numbered badges indicate ground-truth skill types, with the embedded legend mapping each badge to a skill name. Well-separated groups confirm discriminative representation learning.}
\label{fig:tsne}
\end{figure}

Figure~\ref{fig:tsne} visualizes the learned segment embedding space used to inspect whether pseudo-label refinement separates source-domain skill types.

\subsection{Wasserstein Cluster Count Sweep}

Table~\ref{tab:wasserstein_k} reports the unsupervised cluster-count sweep used to choose the $k=8$ clustering analyzed in the main text.

\begin{table}[h!]
\centering
\caption{Wasserstein clustering quality across different numbers of clusters $k$ on IW. We use $k=8$ in the main analysis because it gives the best NMI under the corrected diagonal 2-Wasserstein metric.}
\label{tab:wasserstein_k}
\small
\begin{tabular}{ccc}
\toprule
$k$ & NMI & Purity \\
\midrule
8  & \textbf{0.650} & 0.628 \\
10 & 0.621 & 0.633 \\
12 & 0.564 & 0.633 \\
14 & 0.563 & 0.633 \\
16 & 0.560 & 0.633 \\
\bottomrule
\end{tabular}
\end{table}

\subsection{Action Distribution Diagnostic}

\begin{figure}[h!]
\centering
\includegraphics[width=0.85\textwidth]{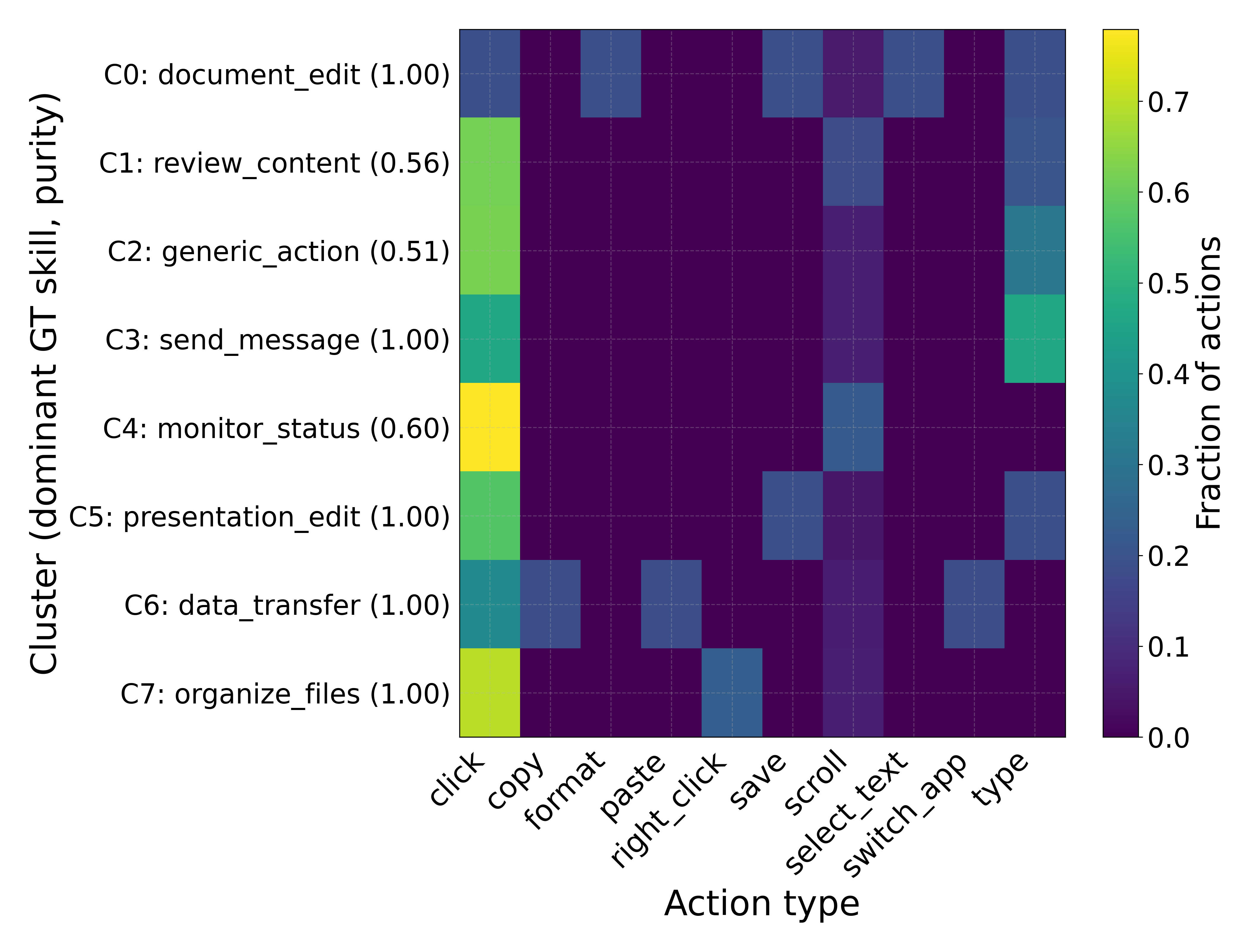}
\caption{Per-cluster action-type distributions for the 8 auto-discovered clusters on IW data. Each row is a cluster, labelled with its dominant ground-truth skill and purity; columns are action types.}
\label{fig:qualitative}
\end{figure}

Figure~\ref{fig:qualitative} gives the action-type distribution behind the qualitative cluster descriptions in Table~\ref{tab:qualitative}.

\subsection{Contrastive Training Curves}
\label{app:training_curves}

\begin{figure}[t]
\centering
\includegraphics[width=0.75\textwidth]{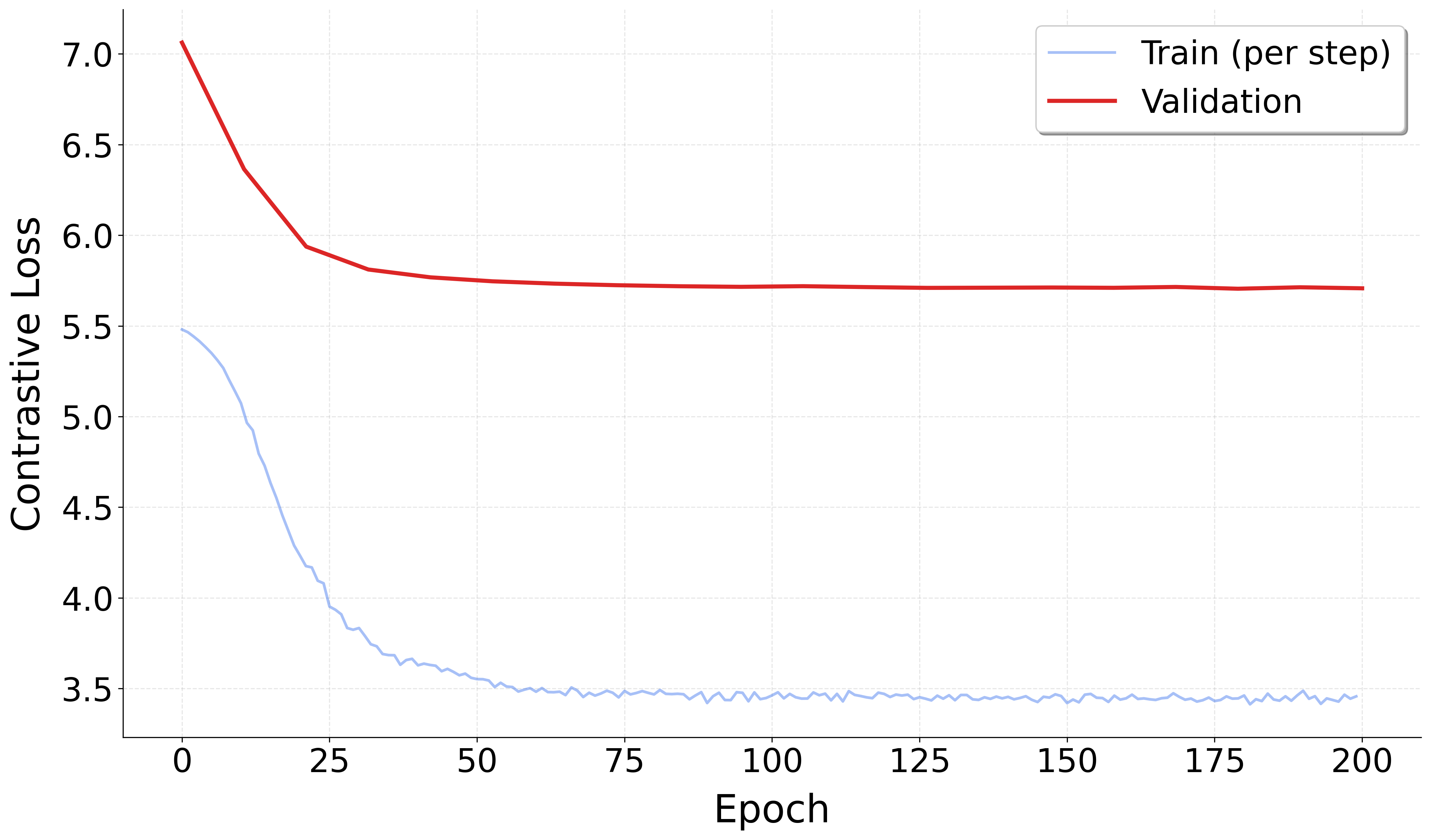}
\caption{Supervised-contrastive training and validation loss over 200 epochs. The horizontal axis denotes epoch. Smooth convergence with stable validation loss indicates no overfitting.}
\label{fig:training}
\end{figure}

Figure~\ref{fig:training} shows the supervised-contrastive optimization trajectory for the skill encoder.

\subsection{Per-Skill Accuracy and Confusion Matrix}
\label{app:perskill}

\begin{figure}[t]
\centering
\includegraphics[width=0.75\textwidth]{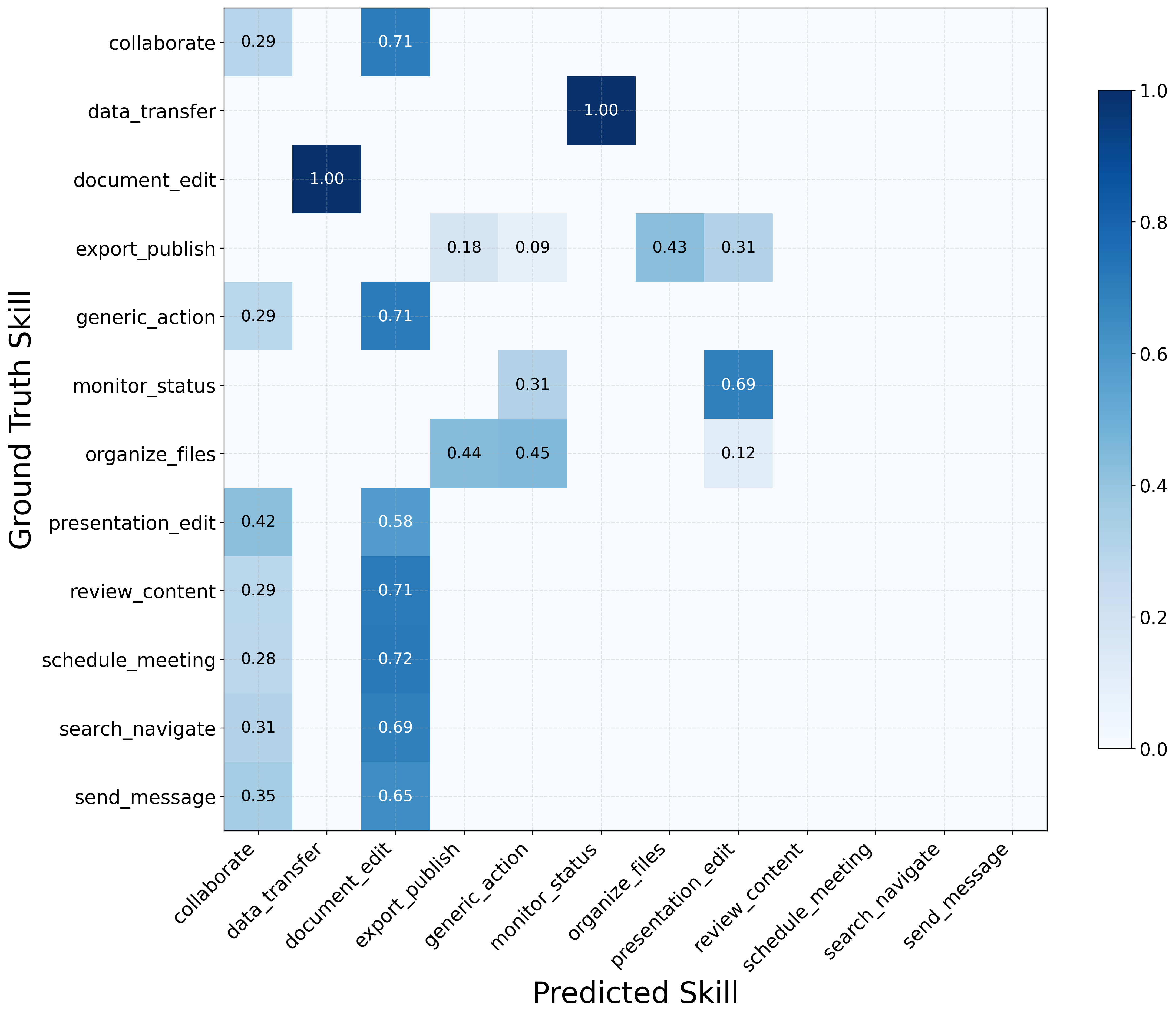}
\caption{Normalized alignment between Wasserstein clustering assignments and ground-truth skill labels on IW data. Rows are discovered clusters and columns are ground-truth skill labels.}
\label{fig:confusion}
\end{figure}

Table~\ref{tab:perskill} reports skill-level accuracy and Figure~\ref{fig:confusion} shows how discovered clusters align with ground-truth skill labels. Here accuracy is the per-step next-skill, or tool-call, classification accuracy: among held-out IW steps whose ground-truth next skill is a given row label, it is the fraction for which the model predicts that same skill label. It is not primitive UI-action accuracy and not exact sequence match.

\begin{table}[t]
\centering
\caption{Per-skill next-skill/tool-call prediction accuracy on IW data, sorted by accuracy. Std. is the binomial standard error over per-step correctness for the single held-out evaluation, and 95\% CI is the Wilson interval.}
\label{tab:perskill}
\footnotesize
\begin{tabular}{lcccc}
\toprule
\textbf{Skill} & \textbf{Acc.} & \textbf{Std.} & \textbf{95\% CI} & \textbf{$n$} \\
\midrule
search\_navigate & 0.941 & 0.033 & [0.841, 0.980] & 51 \\
organize\_files & 0.917 & 0.080 & [0.646, 0.985] & 12 \\
data\_transfer & 0.897 & 0.049 & [0.764, 0.959] & 39 \\
review\_content & 0.893 & 0.058 & [0.728, 0.963] & 28 \\
document\_edit & 0.878 & 0.038 & [0.785, 0.935] & 74 \\
collaborate & 0.831 & 0.044 & [0.727, 0.901] & 71 \\
export\_publish & 0.821 & 0.072 & [0.644, 0.921] & 28 \\
send\_message & 0.809 & 0.048 & [0.700, 0.885] & 68 \\
schedule\_meeting & 0.778 & 0.098 & [0.548, 0.910] & 18 \\
presentation\_edit & 0.556 & 0.166 & [0.267, 0.811] & 9 \\
\bottomrule
\end{tabular}
\end{table}

\subsection{Qualitative Skill Descriptions}
\label{app:qualitative}

For each of the 8 discovered clusters in Table~\ref{tab:qualitative}, we asked Claude Haiku~4.5 to describe the cluster given its top action types, dominant ground-truth skill label, purity, and three nearest-to-centroid exemplar segments. Thus these descriptions are not a blind semantic validation of the clusters: the LLM saw the plurality ground-truth label used in the table. We use them only as concise natural-language interpretations of the same cluster statistics and exemplars reported in Table~\ref{tab:qualitative}. High-purity clusters correspond to cleaner behavioral patterns; low-purity clusters should be read as mixed action motifs rather than clean skills.

\begin{itemize}\setlength\itemsep{0.2em}
    \item C0 (\texttt{monitor\_status}, purity 0.66): Users inspect status-oriented interfaces through repeated click and scroll actions, often drilling into details and returning to nearby controls. The cluster is interpretable as monitoring behavior, but its moderate purity indicates overlap with other navigation-heavy skills.
    \item C1 (\texttt{send\_message}, purity 0.28): Users perform broad click/type routines such as focusing a text field, entering content, and clicking again to submit or advance. Although the plurality label is \texttt{send\_message}, the low purity shows that this cluster mixes several text-entry skills.
    \item C2 (\texttt{document\_edit}, purity 1.00): Users perform document editing and formatting tasks by clicking to position the cursor, selecting and modifying text, applying formatting, and saving changes.
    \item C3 (\texttt{data\_transfer}, purity 1.00): Users copy information from one application and paste it into another by selecting content, copying, switching applications, positioning the cursor, and inserting the data.
    \item C4 (\texttt{organize\_files}, purity 1.00): Users organize files or folders by selecting an item, opening a context menu, and executing an organizational action.
    \item C5 (\texttt{export\_publish}, purity 1.00): Users complete an export or publish workflow through a short sequence of confirmation clicks, consistent with finalizing an already configured artifact.
    \item C6 (\texttt{review\_content}, purity 0.46): Users review or annotate displayed content by scrolling, selecting interface items, entering text, and moving through follow-up clicks. The cluster captures a reusable review-like action motif, but its sub-0.5 purity means it is not a clean one-skill cluster.
    \item C7 (\texttt{presentation\_edit}, purity 1.00): Users edit presentation content by locating content, typing updates, and saving.
\end{itemize}

\section{Generated Skill Specifications}
\label{app:generated_skill_specs}

This section supports the automated \texttt{SKILL.md} analysis in Section~\ref{sec:results_autoskillmd}. The table gives the data-efficiency numbers behind Figure~\ref{fig:autoskillmd_efficiency}; the examples show what the generated skill descriptions look like against hand-written specifications.

\subsection{Data-Efficiency Table}

Table~\ref{tab:autoskillmd} gives the numeric data-efficiency results behind the main-text generated-\texttt{SKILL.md} comparison.

\begin{table}[t]
\centering
\caption{Automated vs.\ hand-crafted \texttt{SKILL.md} generation on IW trajectories. Normalized edit distance is lower better. The hand-crafted baseline is a simple expert-authored transition table. Auto-\texttt{SKILL.md} improves over the hand-crafted table at some sizes but is worse than the Frequency baseline at every evaluated size.}
\label{tab:autoskillmd}
\small
\begin{tabular}{lccccc}
\toprule
\textbf{Method} & \textbf{$N = 100$} & \textbf{$N = 250$} & \textbf{$N = 500$} & \textbf{$N = 1{,}000$} & \textbf{$N = 2{,}000$} \\
\midrule
Frequency & 0.520 & 0.510 & \textbf{0.472} & \textbf{0.494} & 0.485 \\
SKILL.md (hand-crafted) & 0.618 & 0.602 & 0.559 & 0.590 & 0.569 \\
AWM (ground-truth labels) & 0.520 & \textbf{0.510} & 0.477 & 0.496 & \textbf{0.485} \\
\textbf{Auto-SKILL.md (ours)} & \textbf{0.542} & 0.512 & 0.616 & 0.640 & 0.528 \\
\bottomrule
\end{tabular}
\end{table}

\subsection{Hand-Crafted vs. Auto-Generated Skill Examples}
\label{app:skillmd_examples}

The following examples compare hand-crafted and auto-generated \texttt{SKILL.md} descriptions on realistic workflow tasks. They are qualitative examples, not additional metrics. The hand-crafted examples show the kind of detailed manual specification an engineer might write; the auto-generated examples instantiate trajectory-derived reusable structure with the concrete entities, fields, and goals from the task. Across the examples, the main difference is that hand-crafted skills include more expert validation and recovery rules, while auto-generated skills combine reusable action patterns with task-specific context.

\paragraph{Example 1: Service Ticket Update.}
Task: In an IT service-management portal, find an open incident by ticket number, update the assignment group and priority, add a work note, and save the record.

\noindent\begin{minipage}[t]{0.485\linewidth}
{\setlength{\fboxsep}{5pt}
\fcolorbox{handframe}{handbg}{\begin{minipage}{0.91\linewidth}
\scriptsize
\textbf{Hand-crafted \texttt{SKILL.md}}\\
\texttt{update\_service\_ticket}
\begin{itemize}\setlength\itemsep{0.1em}
    \item \textbf{When:} ticket, case, incident, or change request updates.
    \item \textbf{Preconditions:} logged in; search or list filter visible; ticket identifier known.
    \item \textbf{Procedure:} search exact ticket ID; open the matching record; verify record number; locate \texttt{Assignment group}, \texttt{Priority}, \texttt{State}, and \texttt{Work notes}; update requested fields only; save.
    \item \textbf{Validation:} record number unchanged; edited values visible; note appears in activity stream.
    \item \textbf{Recovery:} filter duplicate search results; stop on read-only fields; repair required-field errors before one retry.
\end{itemize}
\end{minipage}}}
\end{minipage}\hfill
\begin{minipage}[t]{0.485\linewidth}
{\setlength{\fboxsep}{5pt}
\fcolorbox{autoframe}{autobg}{\begin{minipage}{0.91\linewidth}
\scriptsize
\textbf{Auto-generated \texttt{SKILL.md}}\\
Cluster C0 mapped to \texttt{record\_field\_update}
\begin{itemize}\setlength\itemsep{0.1em}
    \item \textbf{Observed pattern:} click search/list field $\rightarrow$ type ID $\rightarrow$ open result $\rightarrow$ edit labeled fields $\rightarrow$ click save/update.
    \item \textbf{Task context:} target is an open incident; required fields are assignment group, priority, and work note.
    \item \textbf{Reusable steps:} locate the incident by ticket number; bind task values to nearby labels; enter only requested updates; commit changes.
    \item \textbf{Generated validation:} prefer exact record match; confirm edited labels still show requested values; final save/update terminates the skill.
    \item \textbf{Generated caution:} preserve fields not mentioned in the task.
\end{itemize}
\end{minipage}}}
\end{minipage}

\paragraph{Example 2: Spreadsheet-to-CRM Data Transfer.}
Task: Copy a customer's renewal amount and renewal date from a spreadsheet, switch to a CRM opportunity page, paste the values into the matching fields, and submit the update.

\noindent\begin{minipage}[t]{0.485\linewidth}
{\setlength{\fboxsep}{5pt}
\fcolorbox{handframe}{handbg}{\begin{minipage}{0.91\linewidth}
\scriptsize
\textbf{Hand-crafted \texttt{SKILL.md}}\\
\texttt{transfer\_spreadsheet\_value\_to\_crm}
\begin{itemize}\setlength\itemsep{0.1em}
    \item \textbf{When:} move values from a spreadsheet, table, email, or report into a business system.
    \item \textbf{Preconditions:} source and destination are reachable; source row is identifiable; destination field names are known or inferable.
    \item \textbf{Procedure:} locate source row; select cell; copy; switch to CRM; verify account or opportunity; paste into matching field; repeat for each value; save.
    \item \textbf{Validation:} compare currency symbols, decimals, dates, and units; confirm destination record saved.
    \item \textbf{Recovery:} recopy on clipboard failure; accept destination reformatting only if semantically equivalent; disambiguate duplicate customer rows by renewal period.
\end{itemize}
\end{minipage}}}
\end{minipage}\hfill
\begin{minipage}[t]{0.485\linewidth}
{\setlength{\fboxsep}{5pt}
\fcolorbox{autoframe}{autobg}{\begin{minipage}{0.91\linewidth}
\scriptsize
\textbf{Auto-generated \texttt{SKILL.md}}\\
Cluster C6 mapped to \texttt{data\_transfer}
\begin{itemize}\setlength\itemsep{0.1em}
    \item \textbf{Observed pattern:} click source cell $\rightarrow$ copy value $\rightarrow$ switch app/tab $\rightarrow$ click destination input $\rightarrow$ paste $\rightarrow$ confirm.
    \item \textbf{Task context:} source is the customer's spreadsheet row; values are renewal amount and renewal date; destination is the CRM opportunity page.
    \item \textbf{Reusable steps:} identify row by customer; copy each required value; switch to CRM; bind destination by field label; paste amount and date; submit update.
    \item \textbf{Generated validation:} compare visible pasted strings with source values; allow formatting normalization only after checking semantic equivalence.
    \item \textbf{Generated caution:} app switching is expected and should not reset the task state.
\end{itemize}
\end{minipage}}}
\end{minipage}

\paragraph{Example 3: Shared Drive File Organization.}
Task: In a shared drive, find the latest quarterly budget file, rename it with the approved naming convention, move it into the finance archive folder, and verify that the old copy is no longer in the working folder.

\noindent\begin{minipage}[t]{0.485\linewidth}
{\setlength{\fboxsep}{5pt}
\fcolorbox{handframe}{handbg}{\begin{minipage}{0.91\linewidth}
\scriptsize
\textbf{Hand-crafted \texttt{SKILL.md}}\\
\texttt{organize\_shared\_drive\_file}
\begin{itemize}\setlength\itemsep{0.1em}
    \item \textbf{When:} rename, move, archive, duplicate, or clean up files and folders.
    \item \textbf{Preconditions:} file browser open; source folder accessible; metadata sufficient to identify target file.
    \item \textbf{Procedure:} search or sort folder; identify latest matching file by name, date, and extension; select file; open context menu; rename; confirm; choose move; navigate to archive; confirm move.
    \item \textbf{Validation:} find new filename in destination; check extension and date; verify source folder no longer contains the moved file.
    \item \textbf{Recovery:} disambiguate equally recent files; avoid overwriting filename conflicts; fall back from drag-and-drop to context-menu move.
\end{itemize}
\end{minipage}}}
\end{minipage}\hfill
\begin{minipage}[t]{0.485\linewidth}
{\setlength{\fboxsep}{5pt}
\fcolorbox{autoframe}{autobg}{\begin{minipage}{0.91\linewidth}
\scriptsize
\textbf{Auto-generated \texttt{SKILL.md}}\\
Cluster C7 mapped to \texttt{organize\_files}
\begin{itemize}\setlength\itemsep{0.1em}
    \item \textbf{Observed pattern:} click file-like item $\rightarrow$ open actions menu $\rightarrow$ choose rename/move operation $\rightarrow$ click destination or confirmation.
    \item \textbf{Task context:} target is the latest quarterly budget file; destination is the finance archive folder; operation is rename then move.
    \item \textbf{Reusable steps:} locate latest matching file; select it; rename with requested convention; open move action; choose archive folder; confirm.
    \item \textbf{Generated validation:} confirm the renamed file appears in the destination and no longer appears in the working folder.
    \item \textbf{Generated caution:} menu labels override positional assumptions; confirmation clicks often terminate the skill.
\end{itemize}
\end{minipage}}}
\end{minipage}

\end{document}